\documentclass[journal]{IEEEtran}

\usepackage{cite}
\usepackage{graphicx,amssymb,multirow}
\usepackage{amsmath}
\usepackage{subfig}
\usepackage{epsfig}
\usepackage{algorithm}% http://ctan.org/pkg/algorithms
\usepackage{algorithmicx}
\usepackage{url}
\usepackage{comment}
\usepackage{makecell}
\usepackage{xcolor}
\usepackage{amsfonts}
\usepackage{todonotes}
\usepackage{caption}
\newcommand{\etal}{\emph{et al.}}

\begin{document}
\title{Facial Synthesis from Visual Attributes via Sketch using Multi-Scale Generators}

\author{Xing Di,~\IEEEmembership{Student Member,~IEEE} and 
        Vishal~M.~Patel,~\IEEEmembership{Senior Member,~IEEE}
% <-this % stops a space
\thanks{Xing Di is with the Whiting School of Engineering, Johns Hopkins University, 3400 North Charles Street, Baltimore, MD 21218-2608, e-mail: xing.di@jhu.edu}% <-this % stops a space
\thanks{Vishal M. Patel is with the Whiting School of Engineering, Johns Hopkins University, e-mail: vpatel36@jhu.edu}
\thanks{Manuscript received...}}

\markboth{Journal of \LaTeX\ Class Files,~Vol.~xx, No.~x, Month~2017}%
{Shell \MakeLowercase{\textit{et al.}}: Bare Demo of IEEEtran.cls for IEEE Journals}

\maketitle

\begin{abstract}
Automatic synthesis of faces from visual attributes is an important problem in computer vision and has wide applications in law enforcement and entertainment.   With the advent of  deep generative convolutional neural networks (CNNs), attempts have been made to synthesize face images from attributes and text descriptions.  In this paper, we take a different approach, where we formulate the original problem as a stage-wise learning problem.   We first synthesize the facial sketch corresponding to the visual attributes and then we generate the face image based on the synthesized sketch.   The proposed framework, is based on a combination of  two different Generative Adversarial Networks (GANs) --  (1) a sketch generator network which synthesizes realistic sketch from the input  attributes, and (2) a face generator network which synthesizes facial images from the synthesized sketch images with the help of facial attributes.   Extensive experiments and comparison with recent methods are performed to verify the effectiveness of the proposed attribute-based two-stage face synthesis method.

\end{abstract}

\begin{IEEEkeywords}
face synthesis, visual attributes, generative adversarial networks
\end{IEEEkeywords}

% \fi
\IEEEpeerreviewmaketitle

\section{Introduction}

% motivation

\begin{figure}[t]
	\centering
	\includegraphics[width=0.95\linewidth]{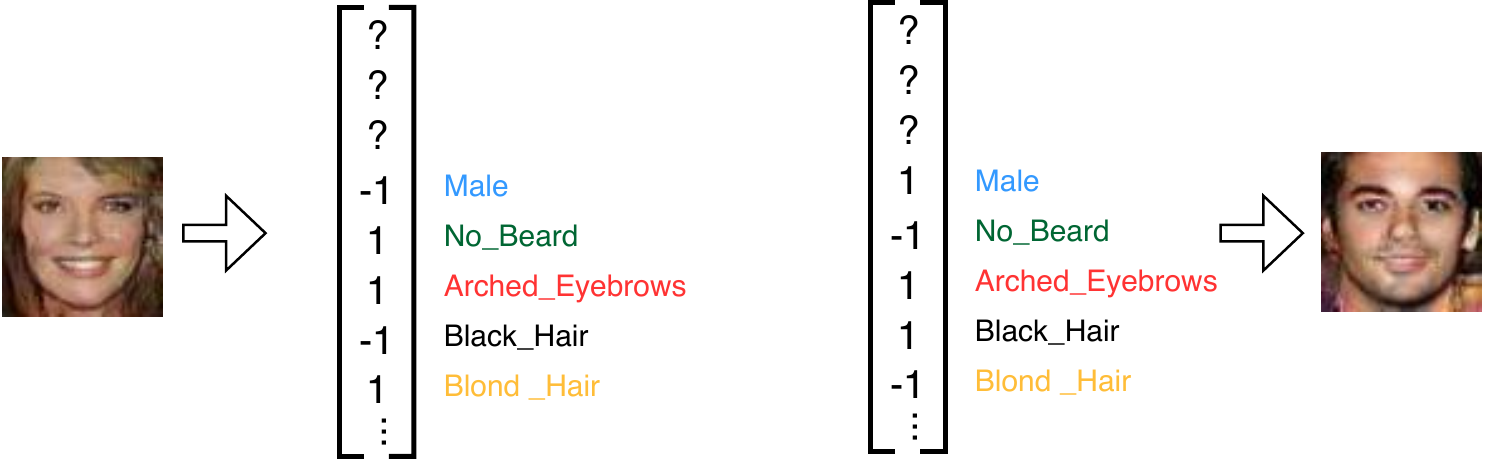}\\
	(a)\hskip100pt (b)
	\caption{Attribute prediction vs. face synthesis from attributes.  (a) Attribute prediction:  given a face image, the goal is to predict the corresponding attributes.  (b) Face synthesis from attributes: given a list of facial attributes, the goal is to generate a face image that satisfies these attributes.}
	\label{fig:att_vs_rec}
\end{figure}

Facial attributes are descriptions or labels that can be given to a face by describing its appearance \cite{kumar_ttributes}.  In the biometrics community, attributes are also referred to as soft-biometrics \cite{softbio}.  Various methods have been developed in the literature for predicting facial attributes from images \cite{DeepAtt,kumar2008facetracer,zhang2014panda,ranjan2019hyperface,rudd2016moon,lu2017fully,gunther2017affact}.  In this work, we aim to tackle the inverse problem of synthesizing faces from their corresponding attributes (see Fig.~\ref{fig:att_vs_rec}). Visual description-based facial synthesis has many applications in law enforcement and entertainment.  For example, visual attributes are commonly used in law enforcement to assist in identifying suspects involved in a crime when no facial image of the suspect is available at the crime scene.  This is commonly done by constructing a composite or forensic sketch of the person based on the visual attributes. 

\begin{figure*}[t]
	\centering
	\includegraphics[width=0.8\linewidth]{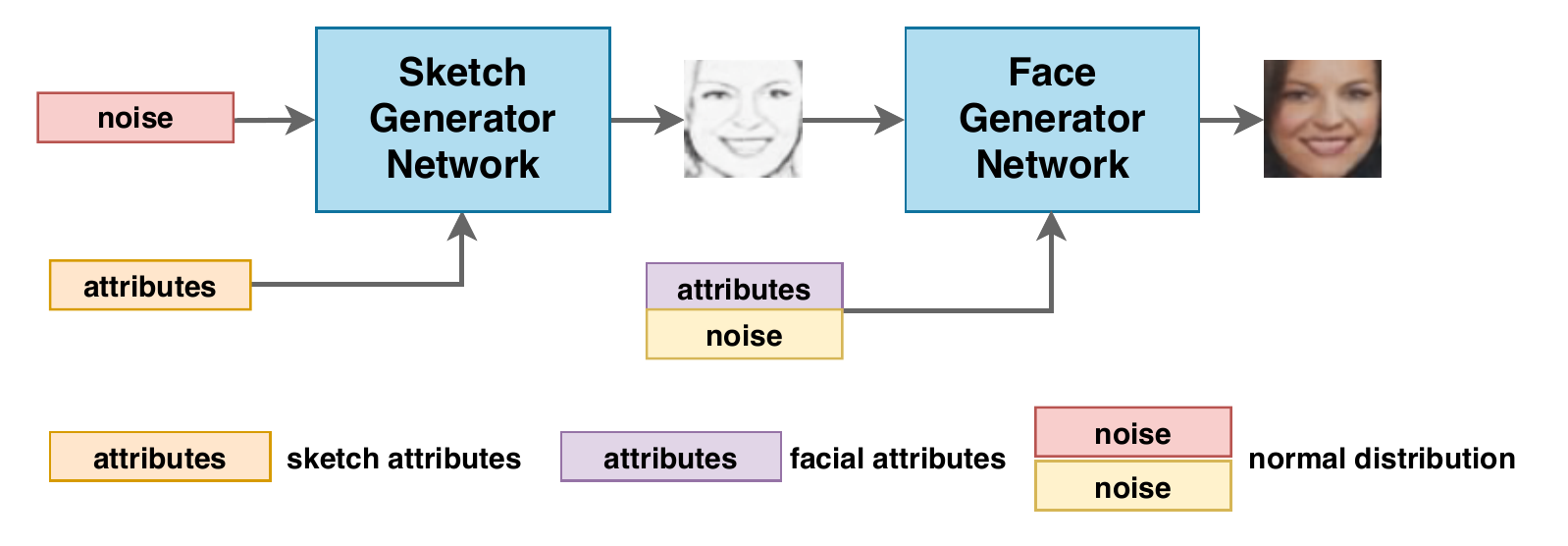}\\
	\caption{An overview of the proposed synthesis method.  Given a noise vector sampled from the normal distribution, the Sketch Generator Network synthesizes sketch image conditioned on the sketch attributes. The synthesized sketch is then given as an input to the Face Generator Network, which outputs the high-quality face images conditioned on the facial attributes. 
	}
	\label{fig:framework}
\end{figure*}

% problem challenges 

Reconstructing an image from attributes or text descriptions is an extremely challenging problem because the model is required to learn the mapping from a semantic abstract space to a complex RGB image space. This task requires the generated images to be not only realistic but also semantically consistent, i.e., the generated face images should preserve the facial structure as well as the content described in attributes.  Several recent works have attempted to solve this problem by using  recently introduced CNN-based generative models such as conditional variational auto-encoder (CVAE) \cite{sohn2015learning,yan2016attribute2image,kingma2013auto} and generative adversarial network (GAN) \cite{goodfellow2014generative,reed2016generative,zhang2016stackgan,zhang2017stackgan++,zhang2018photographic,xu2018attngan}.
For instance, Yan \etal \cite{yan2016attribute2image} proposed a disentangled CVAE-based method for attribute-conditioned image generation.  In a different approach, Reed \etal \cite{reed2016generative} introduced a GAN-based method for synthesizing images from detailed text descriptions.  Similarly, Zhang \etal \cite{zhang2016stackgan} proposed the StackGAN method for synthesizing photo-realistic images from text.
 
It is well-known that CVAE-based methods often generate blurry images due to the injected noise and imperfect element-wise squared error measure used in training \cite{bao2017cvae}.  In contrast, GAN-based methods have shown to  generate high-quality images \cite{goodfellow2014generative}.   In order to synthesize photo-realistic facial images, rather than directly generating an image from attributes, we first synthesize a sketch image corresponding to the attributes and then generate the facial image from the synthesized sketch.  Our approach is motivated by the way forensic sketch artists render the composite sketches of an unknown subject using a number of individually described parts and attributes. Our approach is also inspired by the recent works \cite{Wang_SSGAN2016,sohn2015learning,villegas2017decomposing,Chen_2017_ICCV,Walker_2017_ICCV} that have shown the effectiveness of stage-wise training.

In particular, the proposed framework consists of two stages (see Fig~\ref{fig:framework}) -- sketch generator network and face generator network. Given a noise vector sampled from the normal distribution and sketch attributes, the sketch generator learns to synthesize sketch images.  In the second stage, given the synthesized sketch from the first stage, a different generator network is trained to synthesize high-quality face images with the help of attributes.  In particular,  the attribute augmentation module is adapted from StackGAN \cite{zhang2016stackgan} in both sketch and face generator networks. This module aims to increase the generation diversity by adding redundant information from the standard Gaussian noise. In experiments we observed that, due to the sparsity of  visual attributes, the input attribute values become all zero when the input batch of attributes are all the same. In order to overcome this ``attribute vanishing issue", we replace the batch normalization layers with the conditional batch normalization layers \cite{de2017modulating}. We refer to this module as the attribute augmentation module.  
%Both generators are based on multi-scale networks which help in synthesizing photo realistic images.

To summarize, this paper makes the following contributions:
\begin{itemize}
	\item We formulate the attribute-to-face generation problem as a stage-wise learning problem, i.e. attribute-to-sketch, and  sketch-to-face.  The synthesis networks are based on multi-scale generators with an attribute augmentation module for synthesizing photo-realistic images.
	
	\item For the face generator network, we propose a novel visual attribute conditioned sketch-to-face synthesis network.  The network is composed of an attribute augmentation module and a UNet \cite{ronneberger2015u} shape translation network. With the help of the  attribute-augmentation module, the training stability is  improved and the generators are able to synthesize diverse set of realistic face/sketch images.
	
	\item  Extensive experiments are conducted to demonstrate the effectiveness of the proposed image synthesis method. Furthermore, an ablation study is conducted to demonstrate the improvements obtained by different stages of our framework. 
	
\end{itemize}

Rest of the paper is organized as follows.  In  Section~\ref{sec:related}, we review a few related works.  Details of the proposed facial composite synthesis from visual attribute method are given in Section~\ref{sec:method}.
Experimental results are presented in Section~\ref{sec:expt}, and finally, Section~\ref{sec:con} concludes the paper with a brief summary.

%Code is available at \url{https://github.com/DetionDX/Attribute2Sketch2Face}.

\section{Background and Related Work} \label{sec:related}
Recent advances in deep learning have led to the development of various deep generative models for the problem of text-image synthesis and image-to-image translation \cite{larochelle2011neural}, \cite{kingma2013auto}, \cite{goodfellow2014generative}, \cite{rezende2014stochastic}, \cite{radford2015unsupervised}, \cite{sohn2015learning}, \cite{larsen2015autoencoding}, \cite{denton2015deep}, \cite{dosovitskiy2017learning}, \cite{salimans2016improved}, \cite{metz2016unrolled}, \cite{arjovsky2017towards}, \cite{che2016mode}, \cite{gauthier2014conditional}, \cite{odena2016conditional}.  
Among them, variational autoencoder (VAE) \cite{kingma2013auto,rezende2014stochastic,larsen2015autoencoding}, generative adversarial network (GAN) \cite{goodfellow2014generative,radford2015unsupervised,salimans2016improved,metz2016unrolled,arjovsky2017towards,che2016mode,odena2016conditional}, and Autoregression \cite{larochelle2011neural} are the most widely used approaches. 

VAEs \cite{kingma2013auto,rezende2014stochastic} are powerful generative models that use deep networks to describe distribution of observed and latent variables. A VAE model consists of two parts, with one network encoding a data sample to a latent representation and the other network decoding  latent representation back to data space. VAE regularizes the encoder by imposing a prior over the latent distribution. Conditional VAE (CVAE) \cite{sohn2015learning,yan2016attribute2image} is an extension  of VAE that models latent variables and data, both conditioned on side information such as a part or label of the image. For example, Yan \etal \cite{yan2016attribute2image} proposed a discomposing conditional VAE  (disCVAE) model to synthesize facial image from visual attributes. They took the assumption that a face image could be decomposed into two parts: foreground and background. By taking this assumption, the disCVAE model is able to generate plausible face images with corresponding attributes. However, due to the imperfect element-wise square error measurements, the VAE model usually generates blurry images \cite{bao2017cvae}.

\begin{figure*}[htp!]
	\centering
	\includegraphics[width=.85\linewidth]{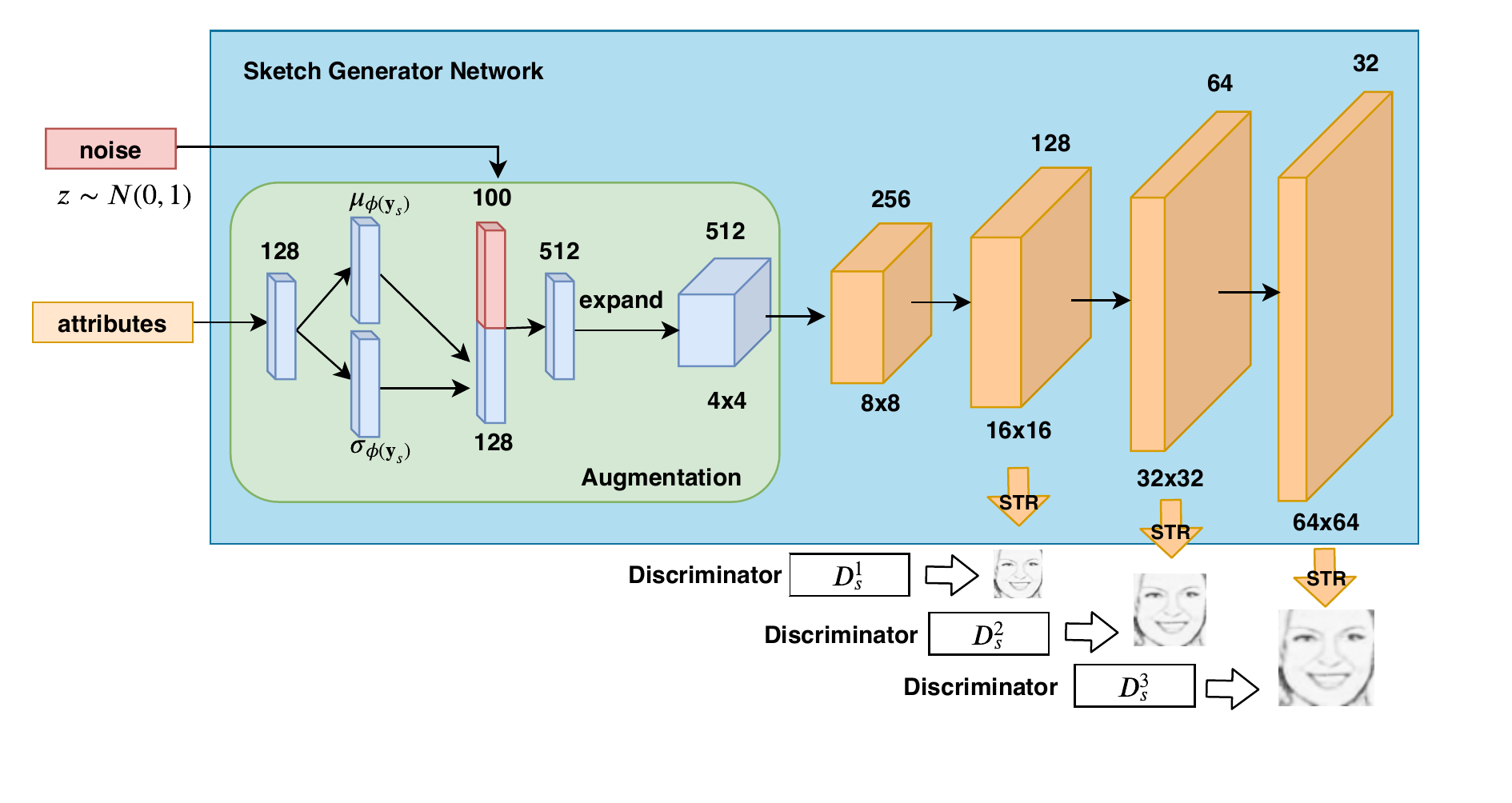}
	\caption{The sketch generator network architecture. The sketch attributes are first augmented by the attribute augmentation module, in which a new latent attribute variable is re-sampled from the estimated latent distribution ($\mu_{\phi(\mathbf{y}_{s})}$ and $\sigma_{\phi(\mathbf{y}_{s})}$)  and concatenated with a noise vector. Then, the remaining up-sample modules (\textcolor{orange}{orange}) aim to generate a series of multi-scale sketches with the augmented sketch attributes.}  
	\label{fig:sketchgeneratornetwork}
\end{figure*}

GANs \cite{goodfellow2014generative} are another class of generative models that are used to synthesize realistic images by effectively  learning the distribution of training images \cite{song2018geometry, lu2018conditional, huang2017beyond}. The goal of GAN is to train a generator $G$, to produce samples from training distribution such that the synthesized samples are indistinguishable from actual distribution by the discriminator, $D$. Conditional GAN is another variant where the generator is conditioned on additional variables such as discrete labels, text  or images. 
The objective function of a conditional  GAN is defined as follows
\begin{equation}\label{eq:conditional GAN loss}
\begin{split}
L_{cGAN}(G,D) = E_{\mathbf{x},\mathbf{y} \sim P_{data} (\mathbf{x},\mathbf{y})}[\log D(\mathbf{x},\mathbf{y})]+ \\
E_{\mathbf{x}\sim P_{data}(\mathbf{x}),\mathbf{z}\sim p_{z}(\mathbf{z})}[\log(1-D(\mathbf{x},G(\mathbf{x},\mathbf{z})))],
\end{split}
\end{equation}
where $\mathbf{z}$ the input noise, $\mathbf{y}$ the output image, and $\mathbf{x}$ the observed image, are sampled from distribution $P_{data} (\mathbf{x},\mathbf{y})$ and they are distinguished by the discriminator, $D$. While for the generated fake $G(\mathbf{x},\mathbf{z})$ sampled from distributions $\mathbf{x}\sim P_{data}(\mathbf{x}),\mathbf{z}\sim p_{z}(\mathbf{z})$ would like to fool $D$. 

Based on these generative models, two common problems have been widely studied by researchers: image-to-image synthesis and text-to-image synthesis.

% image-to-image
\noindent \textbf{Image-to-image synthesis:} One important motivation behind the image-to-image synthesis  problem is to bridge the gap between different image domains. Image-to-image translation models are often built based on the common networks like UNet \cite{ronneberger2015u} and FCN \cite{long2015fully}. Isola \etal \cite{isola2016image} proposed conditional GANs \cite{mirza2014conditional} for several tasks such as labels to street scenes, labels to facades, image colorization, etc. In an another variant, Zhu \etal \cite{zhu2017unpaired} proposed CycleGAN that learns image-to-image translation in an unsupervised fashion. Similarly, Yi \etal \cite{yi2017dualgan} developed an unsupervised dual image-to-image translation model. 

% text-to-image
\noindent \textbf{Text-to-image synthesis:} Isola \etal \cite{isola2016image} proposed Conditional GANs \cite{mirza2014conditional} for several tasks such as labels to street scenes, labels to facades, image colorization, etc.  Built on this, Reed \etal \cite{reed2016generative} proposed a conditional GAN network to generate images conditioned on the text description. Several text-to-image synthesis works have been proposed in the literature that make use of the  multi-scale information  \cite{li2019global,zhang2016stackgan,zhang2017stackgan++,zhang2018photographic,denton2015deep,korshunova2017fast,banerjee2018hallucinating}.   Zhang \etal \cite{zhang2016stackgan} proposed a  two-stage stacked GAN (StackGAN) method which achieves the state-of-the-art image synthesis results. More recently this work was extended in \cite{zhang2017stackgan++} by using  additional losses  and better fine-tuning procedures.  Xu \etal \cite{xu2018attngan} proposed an attention-driven method to improve the synthesis results.  Zhang \etal \cite{zhang2018photographic} (HDGAN) adopted a multi-adversarial loss to improve the synthesis by leveraging more effective image and text information at multi-scale layers.

\section{Proposed Method}\label{sec:method}
In this section, we provide details of the proposed GAN-based attribute to face synthesis method, which consists of two components: sketch generator and face generator. Note that the training phase of our method requires ground truth attributes and the corresponding sketch and face images. Furthermore, the attributes are divided into two separate groups - one corresponding to texture and the other corresponding to color. Since sketch contains no color information, we use only the texture attributes in the first component (i.e. sketch generator) as indicated in  Fig.~\ref{fig:framework}. 

In order to explore the multi-scale information during training, inspired by the previous works \cite{zhang2016stackgan,zhang2017stackgan++,wangfg2018high,zhang2018photographic}, we adopt the idea of hierarchically-integrated multiple discriminators at different layers in our generators. The sketch/face generator network learns the training data distribution from low-resolution to high-resolution. This also helps in improving the training stability of the overall network \cite{karras2017progressive}.

\subsection{Stage 1: Attribute-to-Sketch}
An overview of the sketch generator network architecture is shown in Fig.~\ref{fig:sketchgeneratornetwork}.  Given the sketch attribute vector $\mathbf{y}_{s}$, the goal of the sketch generator network $G_{s}$ is to produce multi-scale sketch outputs as follows
 \begin{equation}\label{multi-scale Gs}
G_{s}(\mathbf{z}_{s}, \mathbf{y}_{s})= \{\mathbf{\hat{x}}_{s}^{1}, \mathbf{\hat{x}}_{s}^{2}, \cdots, \mathbf{\hat{x}}_{s}^{m} \} \triangleq \mathbf{\hat{X}}_{s},
\end{equation}
where $\mathbf{z}_{s}$ is the noise vector sampled from a normal Gaussian distribution, $\{\mathbf{\hat{x}}_{s}^{1}, \mathbf{\hat{x}}_{s}^{2}, \cdots, \mathbf{\hat{x}}_{s}^{m}\}$ are the synthesized sketch images with gradually growing resolutions, and $\mathbf{\hat{x}}_{s}^{m}$ is the final output with the highest resolution.  In order to explore the multi-scale information at different image resolutions, a set of distinct discriminators $D_{s} = \{D_{s}^{1},...,D_{s}^{m}\}$ are implemented for each $\mathbf{\hat{x}}_{s}^{i}, i=1,2,3,\cdots, m$. An example of $3$-scale generator architecture is shown in Fig.~\ref{fig:sketchgeneratornetwork}. It can be observed that the output sketch images are generated from the feature maps with  certain resolutions (width $\times$ height) from different layers of the network.

The generator network consists of three modules: the attribute augmentation module (AA), the up-sample module (UP), and the stretching module (STR). The STR module consists of two $1\times 1$ convolution layers followed by a Tanh layer, which aims to convert the feature map into a 3-channel output image. The UP module consists of an up-sampling layer followed by convolutional, batch normalization, and ReLU layers. Between each UP module, there is  an additional residual block (Res) module \cite{he2016identity,he2016deep}.

The AA module consists of a series of fully-connected neural networks which aim to learn a latent representation of the given visual attribute vector $\mathbf{y}$. During training, we randomly sample a latent variable $\mathbf{\hat{y}}$ from an independent Gaussian distribution $\mathcal{N}(\mu_{\phi(\mathbf{y})}, \sigma_{\phi(\mathbf{y})})$, where the mean $\mu_{\phi(\mathbf{y})}$ and the diagonal covariance matrix $\sigma_{\phi(\mathbf{y})}$ are learned as the functions of visual attributes $\mathbf{y}$. In order to avoid over-fitting, the following KL-divergence regularization term is added during training between the augmented visual attribute distribution and the standard Gaussian distribution
\begin{equation}\label{attribute augmentation}
\mathcal{L}_{aug} = \mathcal{D}_{KL}(\mathcal{N}(\mu_{\phi(\mathbf{y})}, \sigma_{\phi(\mathbf{y})}) \| \mathcal{N}(0, \mathcal{I})),
\end{equation}
where $\mathcal{N}(0, \mathcal{I})$ is the normal Gaussian distribution \cite{zhang2016stackgan,kingma2013auto,larsen2015autoencoding}. Different from previous works \cite{zhang2016stackgan,zhang2017stackgan++}, we replace the traditional batch normalization layers with the conditional batch normalization \cite{de2017modulating} in order to overcome the attribute vanishing problem.

As shown in Fig.~\ref{fig:sketchgeneratornetwork}, the overall sketch generator network architecture is as follows:
\noindent AA(512)-UP(256)-Res(256)-UP(128)Res(128)-UP(64)-Res(64)-UP(32),\\ 
where the number in round bracket indicates the output channel of feature maps. As shown in Fig.~\ref{fig:sketchgeneratornetwork}, the three stretching (STR) modules convert the feature maps into 3-channel output sketch images at different resolutions.

\begin{figure*}[t]
	\centering
	\includegraphics[width=0.85\linewidth]{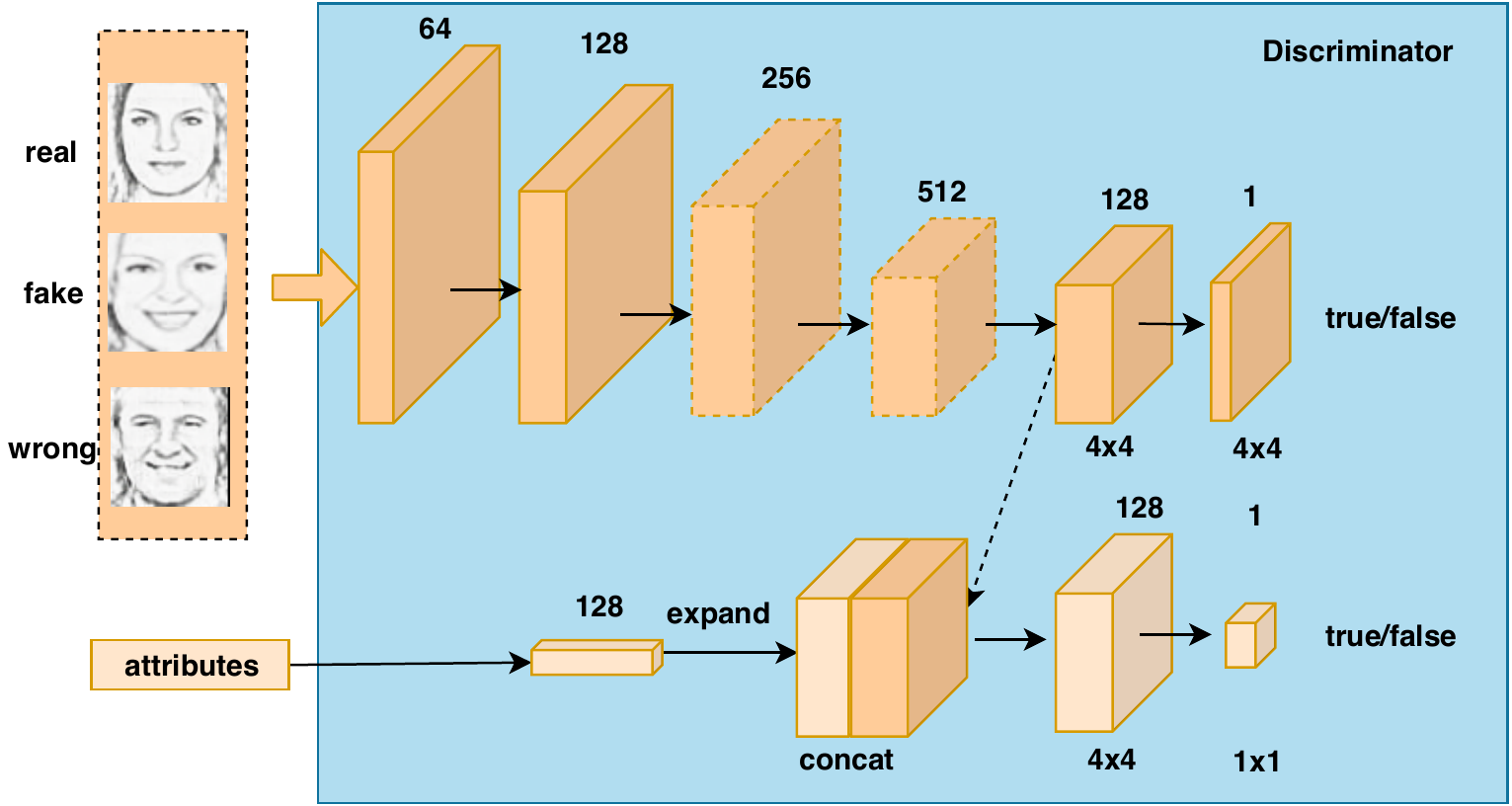}
	\caption{Sketch discriminator at $64\times 64$ resolution scale: given a sketch attribute vector, the discriminator is trained using the triplets: (i) real-sketch and real-sketch-attributes, (ii) synthesized-sketch and real-sketch-attribute, (iii) wrong-sketch (real sketch but mismatching attributes) and same real-sketch-attributes. Note that the convolutional layers with dashed line are removed when training at lower-resolution discriminator. }
	\label{fig:sketchdiscriminatornetwork}
\end{figure*}

\noindent \textbf{Discriminator and Training Loss:} 
  The proposed sketch generator produces multi-scale resolution synthesized sketch images. In order to leverage the hierarchical property of the network, a set of discriminators $D_{s} = \{D_{s}^{1},...,D_{s}^{m}\}$  with similar architectures are designed for each scale.  For a particular scale, the sketch discriminator is developed as shown in Fig.~\ref{fig:sketchdiscriminatornetwork}. In order to learn the discrimination in both image content and semantics, we adopt the triplet matching training strategy  \cite{zhang2017stackgan++,zhang2018photographic,reed2016generative,di2018apgan}. Specifically, given sketch attributes, the discriminator is trained by using the following triplets: (i) real-sketch and real-sketch-attributes, (ii) synthesized-sketch and real-sketch-attributes, and (iii) wrong-sketch (real sketch but mismatching attributes) and same real-sketch-attributes. As shown in Fig.~\ref{fig:sketchdiscriminatornetwork}, two kinds of errors are used to train the discriminator.  They correspond to (i) real/face sketch images, and (ii) sketch images and attributes.

The architecture of the proposed sketch discriminator for $64\times 64$ resolution is shown in Fig.~\ref{fig:sketchdiscriminatornetwork}. This architecture can be easily adapted for other resolution scales by adding/removing appropriate the convolutional layers. As shown in Fig.~\ref{fig:sketchdiscriminatornetwork}, two branches with different losses are used to train the discriminator at a certain resolution scale.  One consists of a series of down-sampling convolutional layers (with filter-size 4, stride 2 and padding size 1) to produce a $4 \times 4$ probability map and classify each location as true or false. The other branch first embeds the sketch attributes to a $128\times4\times4$ feature map and concatenates it with the  feature maps from the first branch. Another two $1\times 1$ convolutional layers are used to fuse the concatenated feature maps to produce $4\times4$ probability maps for classification.  This branch aims to distinguish whether the semantics in sketch images match the sketch attribute or not, through the feedback loss from another $4\times4$ probability map.  

The overall adversarial loss used to train the network is defined as follows:
\begin{equation} \label{eq: multi-scale adversarial loss}
\begin{split}
\mathcal{L}_{s_{Dis}} = \min_{G_{s}} \max_{D_{s}} V(G_{s}, D_{s}, \mathbf{X}_{s}, \mathbf{y}_{s}, \mathbf{z}_{s}) \hspace{30mm}  \\ 
= \sum_{i=1}^{m} \min_{G_{s}} \max_{D_{s}^{i}} (\mathcal{L}_{s_{real}}^{i} + \mathcal{L}_{s_{fake}}^{i} + \mathcal{L}_{s_{wrong}}^{i}), \hspace{15mm}  \\
\mathcal{L}_{s_{real}}^{i} = \mathbb{E}_{\mathbf{x}_{s}^{i}\sim P_{data}(\mathbf{x}_{s}^{i})}[\log D_{s}^{i}(\mathbf{x}_{s}^{i})], \hspace{35mm} \\ + \mathbb{E}_{\mathbf{x}_{s}^{i} \sim P_{data}(\mathbf{x}_{s}^{i}), ,\mathbf{y}_{s} \sim P_{data}(\mathbf{y}_{s})}[\log D_{s}^{i}(\mathbf{x}_{s}^{i},\mathbf{y}_{s})], \hspace{8mm}
\\
\mathcal{L}_{s_{wrong}}^{i} = \mathbb{E}_{\mathbf{x}_{s}^{i'} \sim P_{data}(\mathbf{x}_{s}^{i}), \mathbf{y}_{s}\sim P_{data}(\mathbf{y}_{s})}[\log (1-D_{s}^{i}(\mathbf{x}_{s}^{i'},\mathbf{y}_{s}))], \hspace{0mm}
\\
\mathcal{L}_{s_{fake}}^{i} = \mathbb{E}_{\mathbf{\hat{x}}_{s}^{i} \sim P_{G_{s}(\mathbf{y}_{s},\mathbf{z}_{s})}}[\log (1-D_{s}^{i}(\mathbf{\hat{x}}_{s}^{i}))]  \hspace{26mm}
\\ + \mathbb{E}_{\mathbf{\hat{x}}_{s}^{i} \sim P_{G_{s}(\mathbf{y}_{s},\mathbf{z}_{s})}, \mathbf{y}_{s}\sim P_{data}(\mathbf{y}_{s})}[\log (1-D_{s}^{i}(\mathbf{\hat{x}}_{s}^{i},\mathbf{y}_{s}))],
\end{split}
\end{equation}
where $\mathbf{\hat{x}}_{s}^{i} \sim  P_{G_{s}(\mathbf{y}_{s}, \mathbf{z}_{s})}$ stands for the synthesized (fake) sketch image sampled from the sketch generator at scale $i$,  $\mathbf{x}_{s}^{i}\sim P_{data}(\mathbf{x}_{s}^{i})$ stands for the real sketch image sampled from the sketch image data distribution at scale $i$,  $\mathbf{\hat{x}}_{s}^{i'}$ is the attribute-mismatching sketch image sample at scale $i$, and $\mathbf{y}_{s}$ is the sketch attribute vector. The total objective loss function is given as follows
\begin{equation}\label{generator loss}
\mathcal{L}_{s_{total}} =  \sum_{i=1}^{m} \min_{G_{s}} \max_{D_{s}^{i}} (\mathcal{L}_{s_{real}}^{i} + \mathcal{L}_{s_{fake}}^{i} + \mathcal{L}_{s_{wrong}}^{i}) + \lambda_{s} \mathcal{L}_{s_{aug}},
\end{equation}
where the hyperparameter $\lambda_{s}$ is set equal to 0.01 in our experiments, $ \mathcal{L}_{s_{aug}}$ is the KL-divergence regularization in the AA module with sketch attribute $\mathbf{y}_{s}$ and noise $\mathbf{z}_{s}$ as inputs.

\subsection{Stage 2: Sketch-to-Face} 

%\noindent \textbf{Generator:} 

Given the synthesized sketches $\mathbf{\hat{X}}_{s}$ and the facial attributes $\mathbf{y}_{f}$, the face generator network $G_{f}$ aims to produce multi-scale outputs as follows
\begin{equation}\label{multi-scale Gx}
G_{f}(\mathbf{\hat{X}}_{s}; \mathbf{z}, \mathbf{y}_{f}) = \{\mathbf{\hat{\mathbf{x}}}^{1}_{f}, \mathbf{\hat{\mathbf{x}}}^{2}_{f}, \cdots \mathbf{\hat{\mathbf{x}}}^{m}_{f} \} \triangleq \mathbf{\hat{X}}_{f},
\end{equation}
where $\mathbf{z}$ is noise sampled from a normal Gaussian distribution and $\mathbf{\hat{X}}_{f}$ are the synthesized facial images with gradually growing resolutions. Similar to the sketch generation network, a set of distinct discriminators are designed for each scale. The overall objective is given as follows:
\begin{equation}\label{total min-max face generator}
G_{f}^{\star} , D_{f}^{\star} = arg \min_{G_{f}} \max_{D_{f}} V(G_{f}, D_{f}, \mathbf{X}_{f}; \mathbf{\hat{X}}_{s}, \mathbf{y}_{f}, \mathbf{z}),
\end{equation}
where $D_{f}=\{\mathbb{D}_{1},\cdots, \mathbb{D}_{m} \}$ and $\mathbf{X}_{f}=\{\mathbf{\mathbf{x}}^{1}_{f}, \cdots, \mathbf{\mathbf{x}}^{m}_{f}\}$ denote real training images at multiple scales $1,\cdots, m$.  In order to preserve the geometric structure of the synthesized sketch from the attribute-to-sketch stage, we adopt the skip-connection architecture from UNet related works \cite{di2017gp,di2018apgan,ronneberger2015u}. By using skip-connections, the feature maps from the encoding network are concatenated with the feature maps in the decoding network. This way, the geometric structure of the learned sketch image is  inherited in the synthesized facial image. The proposed method is trained end-to-end.  The lower-resolution outputs fully utilize the top-down knowledge from the discriminators at higher resolutions. Therefore, the synthesized images from different resolutions preserve the geometric structure, which improves the training stability and synthesis quality.

\begin{figure*}
	\centering
	\includegraphics[width=0.85\linewidth]{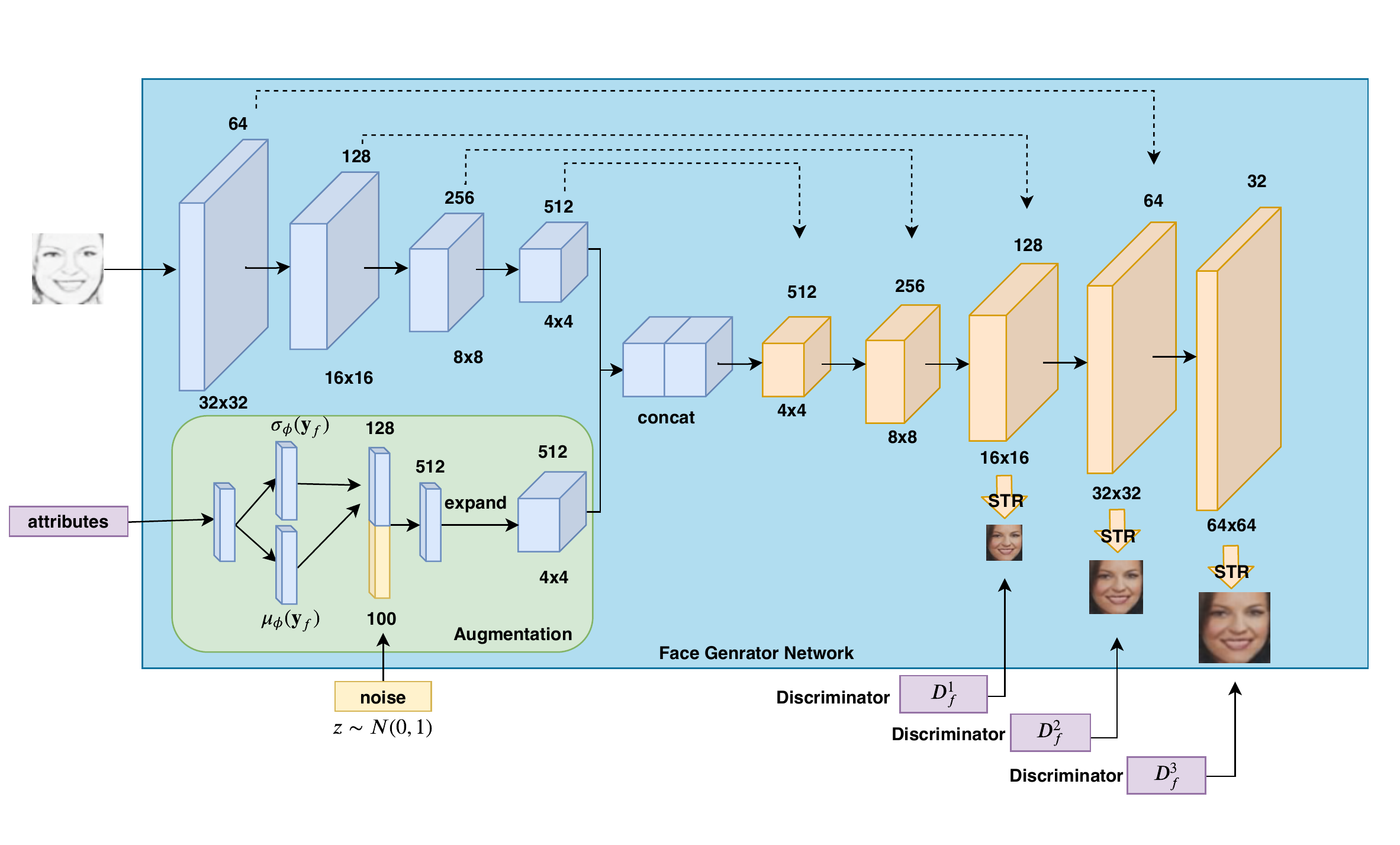}
	\caption{The architecture of Face Generator Network. The facial attributes are first embedded by the attribute augmentation module, similar to the one used in stage 1. The synthesized sketch image is also embedded by \textcolor{blue}{a sequence of down-sample convolutional layers}. These two feature maps are then fused by concatenation. Finally, the fused feature maps are used by the \textcolor{orange}{up-sample module} to synthesize multi-scale face images.}
	\label{fig:facegeneratornetwork}
\end{figure*}

The architecture of the face generator network is shown in Fig.~\ref{fig:facegeneratornetwork}.  The generator consists of four modules: the AA module, the down-sample module (DO), the UP module, and the STR module.  As before, the STR aims convert the feature map into a 3-channel output image.  It consists of two $1\times 1$ convolutional (conv) layers with one Tanh layer.  The UP module consists of an up-sampling layer followed by conv-BN-ReLU layers and an additional residual block \cite{he2016identity,he2016deep} is between each UP module. The DO module consists of a series of conv-BN-ReLU layers. The overall face generator network architecture consists of the following components
\noindent DO(64)-DO(128)-DO(256)-DO(512)-AA(512)-UP(512)-UP(256)-UP(128)-UP(64)-UP(32),
where the number in round brackets indicate the output channel of feature maps.  As shown in Fig.~\ref{fig:facegeneratornetwork}, the three stretching (STR) modules convert the feature maps into 3-channel output face images at different resolutions.

\noindent \textbf{Discriminator and Training Loss:} In the sketch-to-face stage, we use the same architecture of the discriminator as was used in stage 1.  We input the triplets as facial images instead of sketch images and replace the sketch attributes by the facial attributes.  Furthermore, the training loss function is also the same as the one used in the  attribute-to-sketch stage.

\subsection{Testing}
Fig.~\ref{fig:framework} shows the testing phase of the proposed method.  A sketch attribute vector $\mathbf{y}_{s}$ and $\mathbf{z}_{s}$ sampled from a normal Gaussian distribution are first passed through the sketch generator network $G_{s}$ to produce a sketch image.  Then the synthesized sketch image with the highest resolution,  attribute vector $\mathbf{y}_{f}$ and another noise vector $\mathbf{z}_{f}$  are passed through the face generator network to synthesize a face image. In other words, our method takes noise and attribute vectors as inputs and generates high-quality face images via sketch images.  

\section{Experimental Results} \label{sec:expt}
In  this  section,  experimental  settings  and  evaluation  of
the  proposed  method  are  discussed  in  detail.  Results  are  compared  with  several
related generative  models:  disCVAE \cite{yan2016attribute2image}, GAN-INT-CLS  \cite{reed2016generative}, StackGAN \cite{zhang2016stackgan}, Attribute2Sketch2Face \cite{di2017face}, StackGAN++ \cite{zhang2017stackgan++} and HDGAN \cite{zhang2018photographic}. The entire network in Fig.~\ref{fig:framework} is trained end-to-end using Pytorch. When training, the learning rate for the generator and the discriminator in the first stage is set equal to $0.0002$, while the learning rate in the second stage is set equal to $0.0001$.

We conduct experiments using two publicly available datasets: CelebA \cite{liu2015faceattributes}, and deep funneled LFW \cite{Huang2012a}. The CelebA database contains about 202,599 face images, 10,177 different identities and 40 binary attributes for each face image.  The deep funneled LFW database contains about 13,233 images, 5,749 different identities and 40 binary attributes for each face image which are from the LFWA dataset \cite{liu2015faceattributes}. 

\begin{figure}[htp!]
	\centering
	\includegraphics[width=1\linewidth]{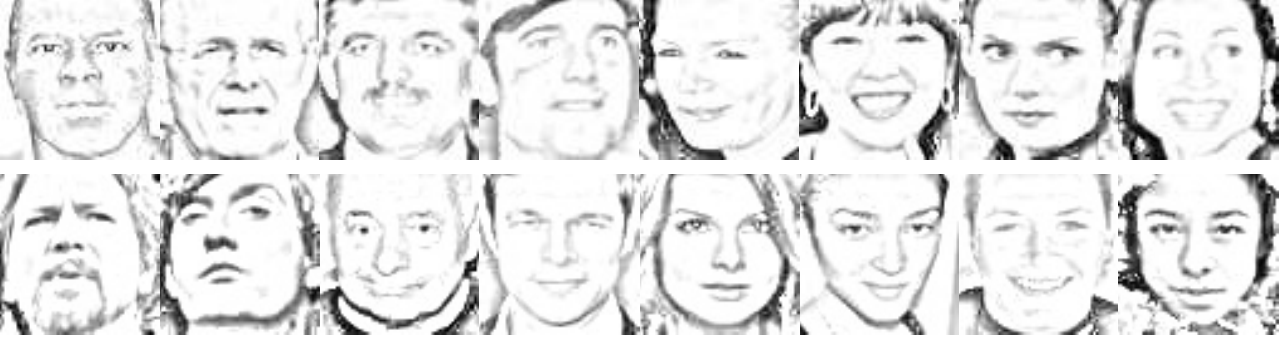}
	\caption{Sketch images sampled from the LFW and the CelebA datasets are shown in row 1 and row 2 respectively.}
	\label{fig:sketch_example}
\end{figure}

Note that the training part of our network requires original face images and the corresponding sketch images as well as the corresponding list of visual attributes.  The CelebA and the deep funneled LFW datasets consist of both the original images and the corresponding attributes.
%while the CUHK dataset consists of face-sketch image pairs.   
To generate the missing sketch images in the CelebA and the deep funneled LFW datasets, we use a public pencil-sketch synthesis method \footnote{http://www.askaswiss.com/2016/01/how-to-create-pencil-sketch-opencv-python.html} to generate the sketch images from the face images.   
Fig.~\ref{fig:sketch_example} shows some sample generated sketch images from  the CelebA and the deep funneled LFW datasets. 

The MTCNN method \cite{zhang2016joint} was used to detect and crop faces from the original images.  The detected faces were scaled to the size of $64\times 64$. 
%\textbf{Since some attributes from the original list of 40 attributes are limited shown in the images, we sorted the attributes by their show-up times and 23-top attributes are selected in this work.} 
Since many attributes from the original list of 40 attributes were not significantly informative, we selected 23 most useful attributes for our problem.  Furthermore, the selected attributes were further divided into 17 texture and 6 color attributes as shown in Table~\ref{tab:fine-grained attributes}. During experiments, the texture attributes were used to train the sketch generator network while all 23 attributes were used to train the face generator network. 

\begin{table}[htp!]
	\centering
	\caption{List of fine-grained texture and color attributes.}\label{tab:fine-grained attributes} 
	\begin{tabular}{|c|c|}
		\hline  Texture &  \makecell{5\_o\_Clock\_Shadow, Arched\_Eyebrows,  Bags\_Under\_Eyes, \\ Bald, Bangs, Big\_Lips, Big\_Nose, Bushy\_Eyebrows,\\ Chubby, Eyeglasses, Male, Mouth\_Slightly\_Open,\\ Narrow\_Eyes,  No\_Beard, Oval\_Face,  Smiling, Young} \\ 
		\hline  Color& \makecell{Black\_Hair, Blond\_Hair, Brown\_Hair, \\Gray\_Hair, Pale\_Skin, Rosy\_Cheeks} \\ 
		\hline 
	\end{tabular}
\end{table}

\subsection{CelebA Dataset Results}

\begin{figure*}[htp!]
\centering
\includegraphics[width=0.85\linewidth]{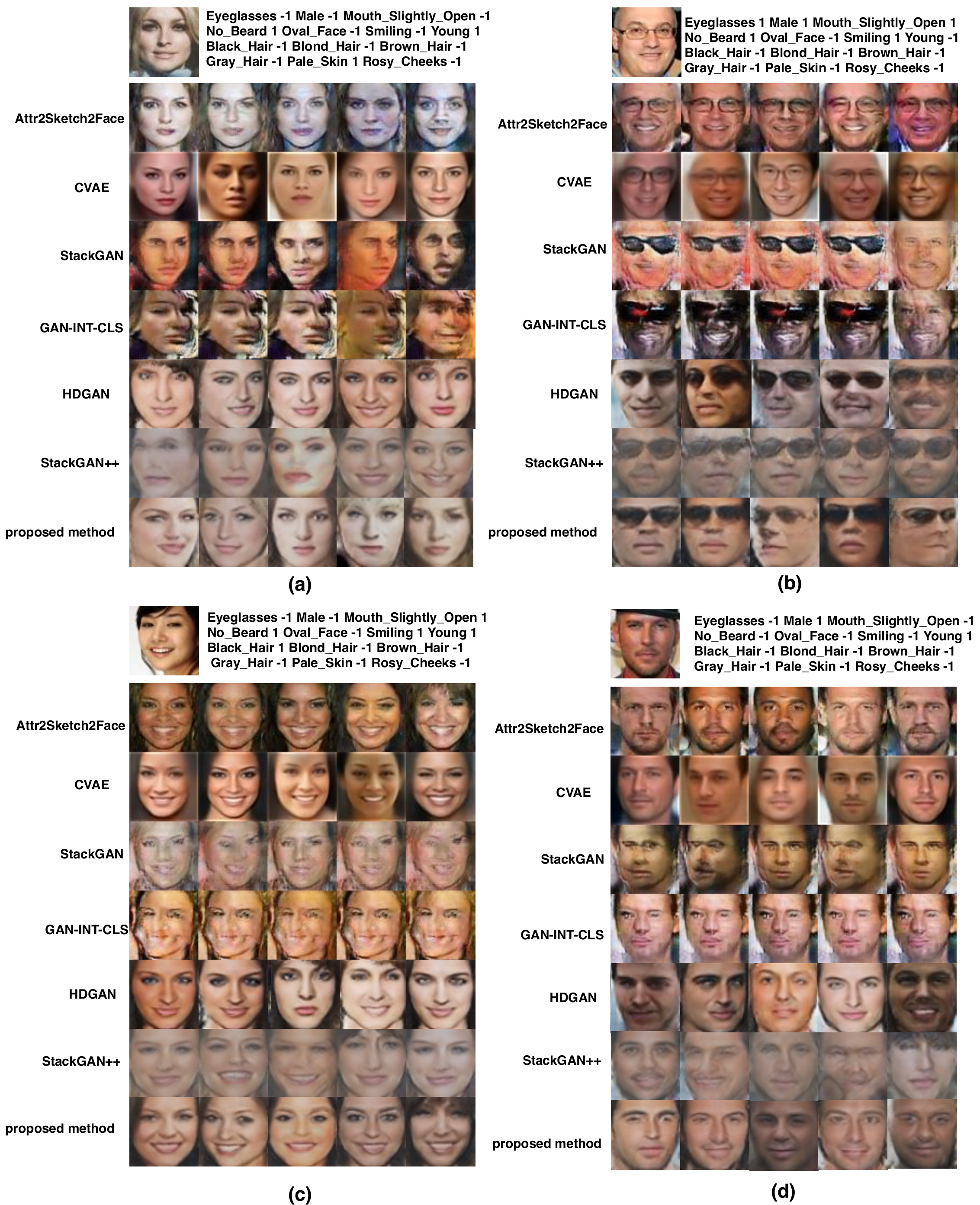}
\caption{Image generation results on the CelebaA dataset. First row of each sub-figure shows the reference image and its corresponding attributes.  The images generated by different methods are shown in different rows.}
\label{fig:CelebA_result}
\end{figure*}

The CelebA dataset \cite{liu2015faceattributes} consists of 162,770 training samples, 19,867 validation samples and 19,962 test samples. We combine the training and validation splits together to train our models. After detection and alignment, we obtain 182,468 samples which we use for training our proposed model. During training, we use the batch size of 40.  The ADAM algorithm \cite{adam_opt} with learning rate of 0.0002 is used to train the network.  In total, 20 training epoch are used during training and the initial learning rate is frozen for the first 10 epochs.  For the next 10 epochs, we let it drop by 0.1 of the initial value after every epoch. The latent feature dimension for sketch/facial attribute is set equal to 128.  The noise vector dimension is set equal to 100. Three scales ($16 \times 16$, $32\times 32$, and $64\times 64$) are used in our multi-scale network.

Sample image generation results corresponding to different methods from the CelebA are shown in Fig.~\ref{fig:CelebA_result}. For fair comparison with those stage-wise training algorithm, we adopt StackGAN \cite{zhang2016stackgan} network to two scale resolution $32\times 32$ and $64\times 64$. Moreover, we adopt StackGAN++ \cite{zhang2017stackgan++} HDGAN \cite{zhang2018photographic} in the same resolution scales: $16\times 16$, $32\times 32$, and $64\times 64$. Note that these results are obtained by inputting a certain attribute vector along with random noise. As can be seen from this figure, GAN-INT-CLS and StackGAN methods easily meet the modal collapse issue in this problem. During training, the generator learns to generate a limited number (1 or 2) of image samples corresponding to a certain list of attributes. This synthesized results are good enough to fool the discriminator. Thus, the generator and discriminator networks do not get optimized properly. The disCVAE method is able to reconstruct the images without model collapse but they are blurry due to the imperfect $L_{2}$ measure in the Gaussian distribution loss. In addition, some of the attributes are difficult to see in the reconstructions corresponding to the disCVAE method, such as the hair color. This is because of the imperfect latent embedding by the variational bound limitation in VAE. Also, the other Attribute2Sketch2Face is able to generate realistic results, but the image quality is slightly inferior.  The recent state-of-art text-to-image synthesis approaches (stackGAN++ and HDGAN) generate plausible facial images from visual attributes.  However, the generated facial images do not always preserve the corresponding attributes very well. Compared with all the baselines, the proposed method not only generates realistic facial images but also preserves the attributes better than the others.  We believe that this is mainly due to the way we appraoch the attribute-to-face synthesis apporach by decomposing it into two problems, attribute-to-sketch and sketch-to-face.  By factoring the original problem into two separate problems, the model at each stage learns better conditional data distribution.  Furthermore, the use of multi-scale generators in the proposed GANs also help in improving the performance of our method.

\subsection{LFW Dataset Results}

\begin{figure*}[htp!]
	\centering
	\includegraphics[width=0.85\linewidth]{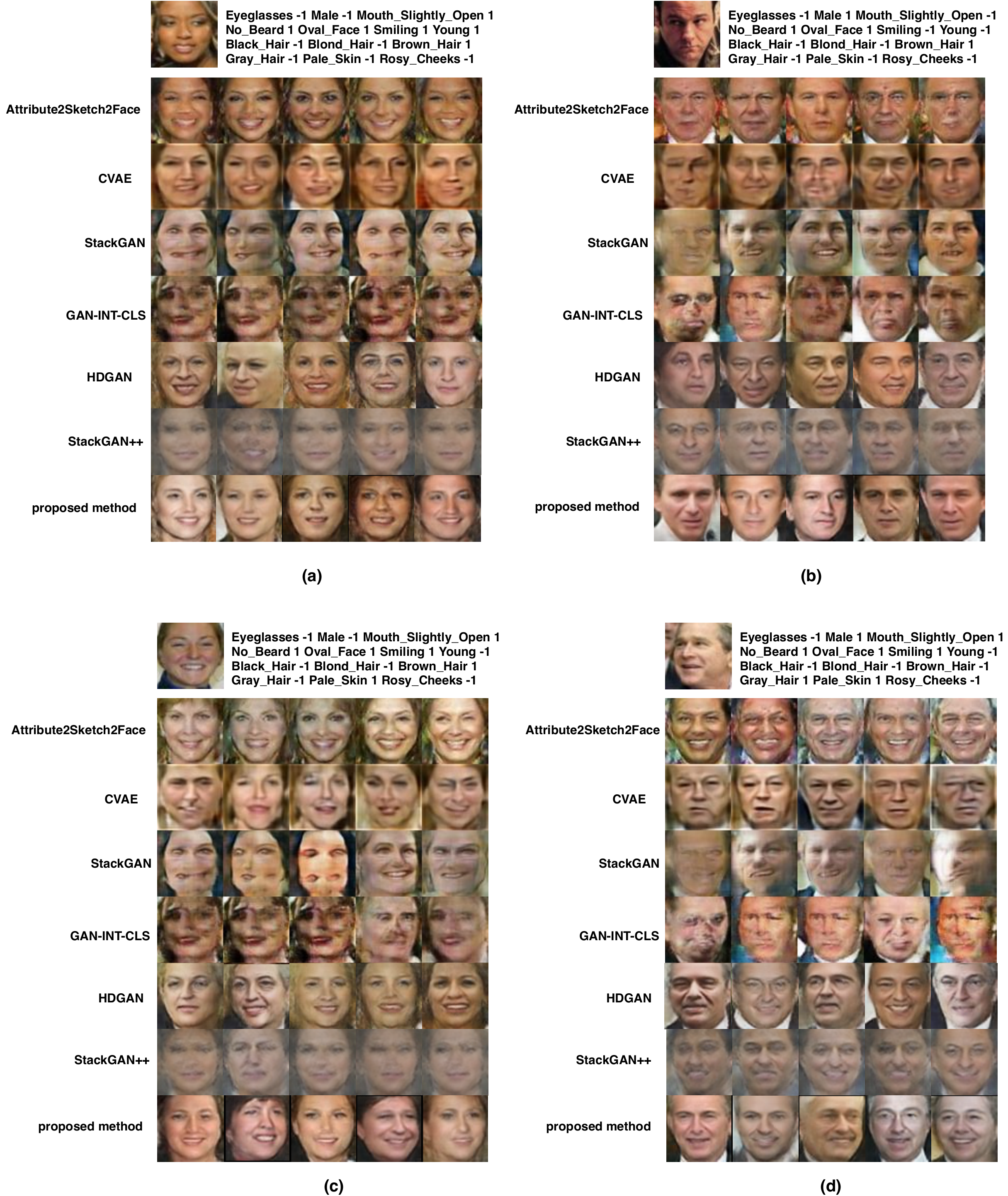}
	\caption{Image generation results on the LFWA dataset. First row of each sub-figure shows the reference image and its corresponding attributes.  The images generated by different methods are shown in different rows.}
	\label{fig:LFW_result}
\end{figure*}

Images in the LFWA dataset come from the LFW dataset \cite{Huang2012a}, \cite{LFWTech}, and the corresponding attributes come from \cite{liu2015faceattributes}. This dataset contains the same 40 binary attributes as in the CelebA dataset. After pre-processing, the training and testing subsets contain 6,263 and 6,880 samples, respectively. We use all the training splits to train our model. The ADAM algorithm \cite{adam_opt} with learning rate of 0.0002 is used for both generators and discriminators. The initial learning rate is frozen in the first 100 epochs and is then dropped by 0.01 for the remaining 100 epochs. All the other experimental settings are the same as the ones used in with CelebA dataset.

Sample results corresponding to different methods on the LFWA dataset are shown  in Fig.~\ref{fig:LFW_result}. For fair comparison, the multi-scale resolution settings are the same as used with the experiments on the CelebA dataset. In particular, we use $32\times 32$ and $64\times 64$ resolution scales for StackGAN \cite{zhang2016stackgan} training and $16\times 16$, $32\times 32$ and $64\times 64$ multiple resolution scales for HDGAN \cite{zhang2018photographic} and StackGAN++ \cite{zhang2017stackgan++} as well as our proposed method. The disCVAE method produces reconstructions which are blurry.  Previous conditional GAN-based approaches such as  GAN-INT-CLS \cite{reed2016generative} and  StackGAN \cite{zhang2016stackgan} also  produce poor quality results due to the model collapse during training. Recent StackGAN++ and HDGAN works generate plausible facial images (HDGAN is better at the color diversity). The previous work Attribute2Sketch2Face, which is a combination of CVAE and GAN, is also able to generate facial images with corresponding attributes. However, the proposed method is able to reconstruct high-quality attribute-preserved face images better than the previous approaches.   

\subsection{CelebA-HQ Dataset Results}
\begin{figure*}[!htb]
	\centering
	\begin{minipage}{0.95\textwidth}
		\centering
		\includegraphics[width=0.95\linewidth]{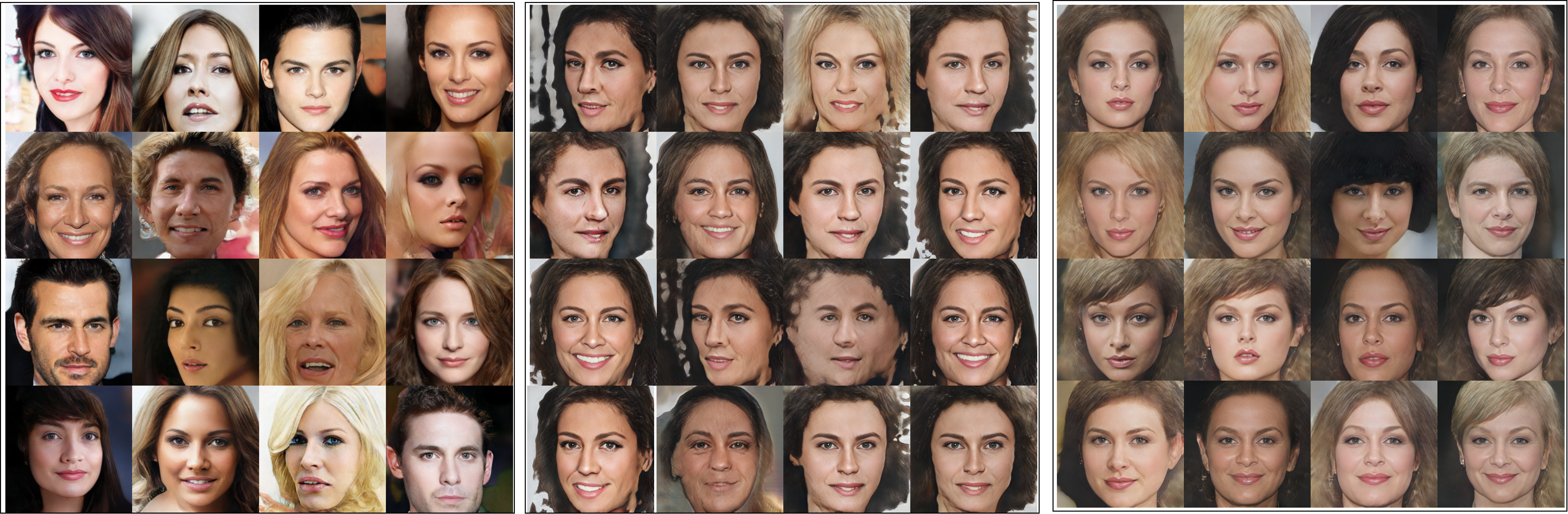} \\
		StackGAN++ \cite{zhang2017stackgan++} \hspace{35mm} HDGAN \cite{zhang2018photographic} \hspace{35mm} Proposed method 
		
		\caption{Image synthesis results on $256\times 256$ resolution.  The attributes used to generate these images are: Eyeglasses -1 Male -1 Mouth\_Slightly\_Open -1 No\_Beard 1 Oval\_Face -1 Smiling -1 Young 1	Black\_Hair -1 Blond\_Hair -1 Brown\_Hair -1 Gray\_Hair -1 Pale\_Skin 1 Rosy\_Cheeks -1.}
		\label{fig:celebahq_comparison256}
	\end{minipage}

\end{figure*}

In order to demonstrate how our proposed method works on high-resolution images, we also conduct an experiment using a recent proposed CelebA-HQ dataset.  The CelebA-HQ dataset \cite{karras2017progressive} is a high-quality version of the CelebA dataset, which consists of 30,000 images with $1024\times 1024$ resolution. Due to GPU and memory limitations, we conduct experiments on $256\times 256$ resolution images and compare the performance with StackGAN++ \cite{zhang2017stackgan++} and HDGAN \cite{zhang2018photographic}. The reason why we chose these two baselines is due to their capability to deal with high resolution images.  Sample results are shown in Figure~\ref{fig:celebahq_comparison256}.

For fair comparison, we set the number of resolution scale $s=3$ for all methods. In order to adopt our method to the high-resolution dataset, we follow the strategy that removing/adding the number of UP/DO block (as defined in Section~\ref{sec:method}) in the generator and the discriminator. In particular, we set the STR modules at resolution $64 \times 64$, $128 \times 128$ and $256 \times 256$ respectively. In experiments, the batch-size is set equal to 16 for our proposed method, which is smaller than StackGAN++ and HDGAN, which are set equal to 24, due to the GPU memory limitations. Also, when training on this dataset, we train the sketch generator first and then use the pre-trained model for training the face generator.

As can be seen from Figure~\ref{fig:celebahq_comparison256}, our proposed method can synthesize photo-realistic images on high-resolution images as well. Moreover, when we compare the attributes from the synthesized images with the given attributes, we can observe that our method preserves the attributes better than the other methods.  Quantitative comparisons in terms of the FID scores also show that the proposed method performs favorably compared to StackGAN++ and HDGAN.  In addition, comarison of our method in Table~\ref{tab:fid celebahq} with only a single scale shows the significance of our multi-scale network.   

\begin{table}
	\centering
\begin{tabular}{|c|c|}
	\hline 
	Methods & FID score   \\ 
	\hline 
	HDGAN \cite{zhang2018photographic} & 114.912   \\ 
	\hline
	StackGAN++\cite{zhang2017stackgan++} & 35.988  \\
	\hline 
	Single-scale (proposed method) & 37.381 \\
	\hline
	Proposed method & 30.566 \\
	\hline 
\end{tabular}
	\caption{Quantitative results (FID scores) corresponding to different methods on the CelebA-HQ dataset.}
	\label{tab:fid celebahq}
\end{table}

\subsection{Face Synthesis}
\begin{figure*}[htp!]
	\centering
	\includegraphics[width=0.85\linewidth]{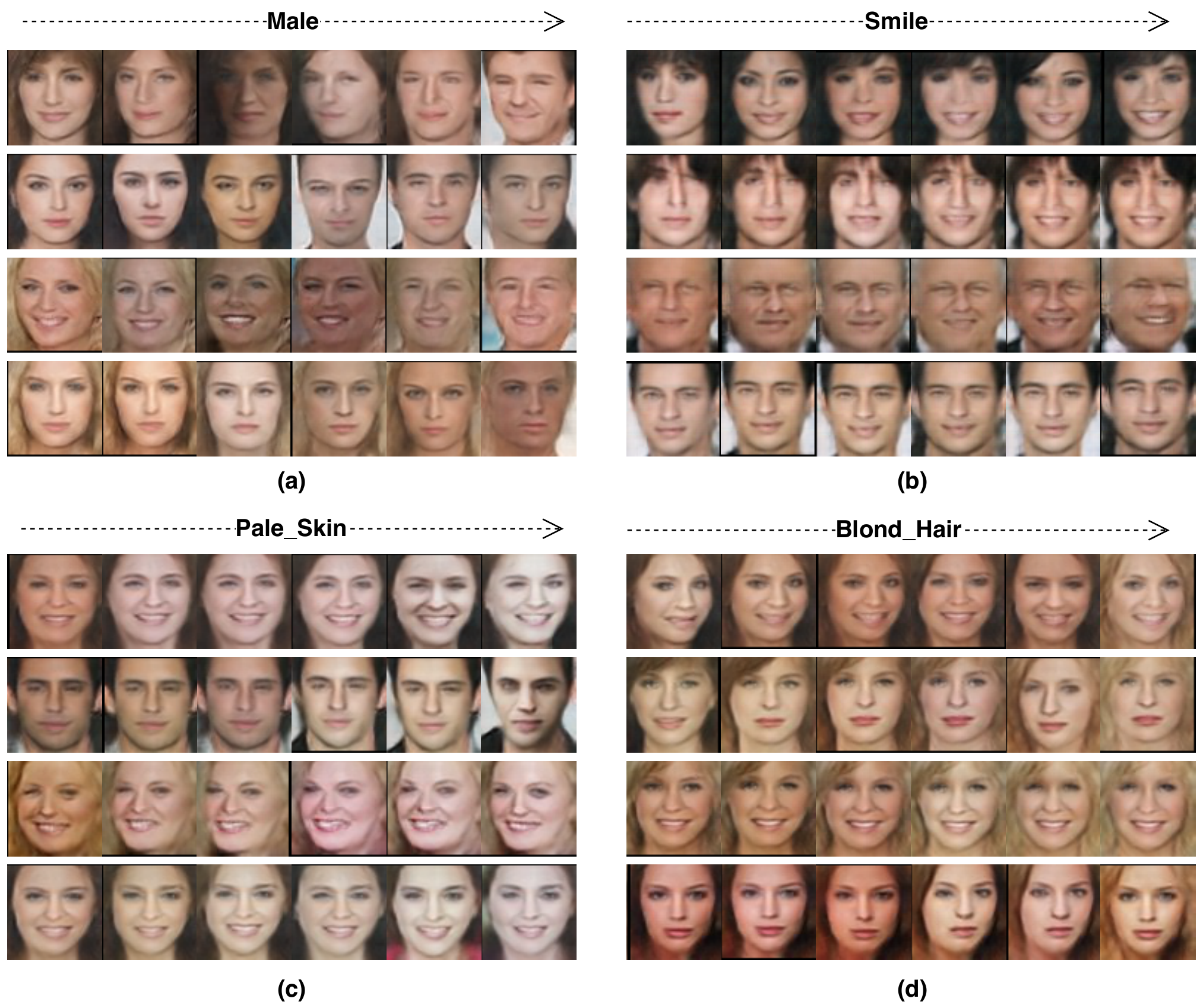}
	\caption{Facial image progressive synthesis on CelebA when attributes are changed. These progressive changes are based on one certain attribute manipulating while the others are keep frozen.  (a) Male. (b) Smile. (c) Original skin tone to pale skin tone. (d) Original hair color to black hair color.}
	\label{fig:CelebA_progression}
\end{figure*}

\begin{figure*}[htp!]
	\centering
	\includegraphics[width=0.85\linewidth]{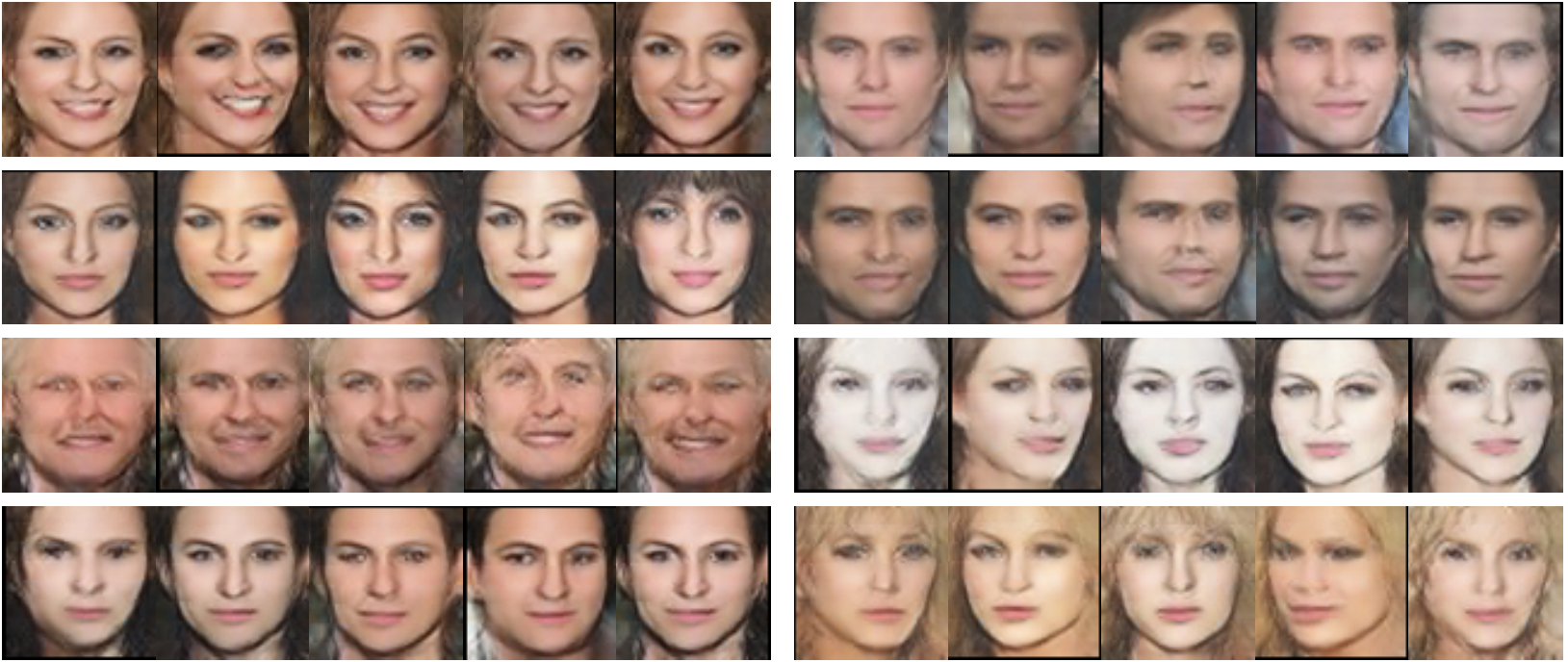}
	\caption{Facial image synthesis sampled when attributes are kept frozen while the noise vector is changed. Note that the identity, pose, or facial shape changes as we vary the noise vector but he attributes stay the same on the synthesized images.}
	\label{fig:random_noise}
\end{figure*}

\begin{comment}
\begin{table*}[htp!]
	\centering
	\caption{Quantitative results corresponding to different methods.The Inception Score and Attribute $L_{2}$ measure are used to compare the performance of different methods.}
	\label{tab: quantitative result}
	\begin{tabular}{|c|c|c|c|c|c|c|}
		\hline
		& \multicolumn{3}{c|}{LFW}          & \multicolumn{3}{c|}{CelebA}       \\ \hline
		Baselines & FID Score & Inception Score & Attribute $L_{2}$ & FID Score & Inception Score & Attribute $L_{2}$ \\ \hline
		GAN-INT-CLS \cite{reed2016generative}    & $85.811$    &$1.510 \pm 0.020$&$0.093 \pm 0.027$& $92.793$ & $1.486 \pm 0.016$&$0.104 \pm 0.024$           \\ \hline
		disCVAE \cite{yan2016attribute2image}&  $103.855$   &$1.275 \pm 0.005$&$0.086\pm 0.019$& $91.012$ &$1.482 \pm 0.017$&$0.080\pm 0.042$\\ \hline
		StackGAN \cite{zhang2016stackgan}&  $70.379$   &$1.517 \pm 0.014$ &$0.085\pm 0.029$&  $63.816$   &$1.589 \pm 0.018$&$0.091 \mp 0.021$ \\ \hline
		StackGAN-v2\cite{zhang2017stackgan++}&   $50.360$  &$2.011 \pm 0.027$&           &   $49.889$  &$2.105 \pm 0.019$&           \\ \hline
		HDGAN \cite{zhang2018photographic}&  $48.930$   &$2.117 \pm 0.027$&           &  $43.206$   &$2.357 \pm 0.037$&           \\ \hline
		Attribute2Sketch2Face \cite{di2017face} &  $60.487$   & $1.637 \pm 0.025$ &$0.059 \pm 0.034$&  $58.896$   & $1.657 \pm 0.011$ &$0.067 \pm 0.022$\\ \hline
		Attribute2Sketch2Face-v2 &  $\mathbf{43.712}$   & $\mathbf{2.353 \pm 0.029}$  &           &  $\mathbf{33.497}$   &   $\mathbf{2.515 \pm 0.017}$ &           \\ \hline
	\end{tabular}
\end{table*}
\end{comment}

\begin{table*}[htp!]
	\centering
	\caption{Quantitative results corresponding to different methods.The FID score and Attribute $L_{2}$ measure are used to compare the performance of different methods.}
	\label{tab: quantitative result}
	\begin{tabular}{|c|c|c|c|c|}
		\hline
		& \multicolumn{2}{c|}{LFW}          & \multicolumn{2}{c|}{CelebA}       \\ \hline
		Baselines & FID Score  & Attribute $L_{2}$ & FID Score  & Attribute $L_{2}$ \\ \hline
		GAN-INT-CLS \cite{reed2016generative}    & $85.811$   &$0.093 \pm 0.027$& $92.793$ &$0.104 \pm 0.024$           \\ \hline
		disCVAE \cite{yan2016attribute2image}&  $103.855$   &$0.086\pm 0.019$& $91.012$ &$0.080\pm 0.042$\\ \hline
		StackGAN \cite{zhang2016stackgan}&  $70.379$    &$0.085\pm 0.029$&  $63.816$   &$0.091 \pm 0.021$ \\ \hline
		StackGAN++\cite{zhang2017stackgan++}&   $50.360$  & $0.059 \pm 0.026$          &   $49.889$  &     $0.061 \pm 0.026$      \\ \hline
		HDGAN \cite{zhang2018photographic}&  $48.930$   &  $0.053 \pm 0.023$         &  $43.206$   &     $0.056 \pm 0.020$     \\ \hline
		Attribute2Sketch2Face \cite{di2017face} &  $60.487$   &$0.059 \pm 0.034$&  $58.896$   & $0.067 \pm 0.022$\\ \hline
		Attribute2Sketch2Face-v2 (proposed method) &  $\mathbf{43.712}$   &  $\mathbf{0.048 \pm 0.020}$         &  $\mathbf{33.497}$   &  $\mathbf{0.051 \pm 0.019}$            \\ \hline
	\end{tabular}
\end{table*}

In this section, we show the image synthesis capability of our network by manipulating the input attribute and noise vectors.  Note that, the testing phase of our network takes attribute vector and noise as inputs and produces synthesized face as the output. In the first set of experiments with image synthesis, we keep the random noise vector frozen and change the weight of a particular attribute as follows: $[-1, -0.1, 0.1, 0.4, 0.7, 1]$.  The corresponding results on the CelebA dataset are shown in Fig.~\ref{fig:CelebA_progression}.  From this figure, we can see that when we give higher weights to a certain attribute, the corresponding appearance changes.  For example, one can 
synthesize an image with a different gender by changing the weights corresponding to the gender attribute as shown in Fig.~\ref{fig:CelebA_progression}(a).  Each row shows the progression of gender change as the attribute weights are changed from -1 to 1 as described above.  Similarly, figures (b), (c) and (d) show the synthesis results when a neutral face image is transformed into a smily face image, skin tones are changed to pale skin tone, and hair colors are changed to black, respectively.    It is interesting to see that when the attribute weights other than the gender attribute are changed, the identity of the person does not change too much. 

In the second set of experiments, we keep the input attribute vector frozen but now change the noise vector by inputing different  realizations of the standard Gaussian.  Sample results corresponding to this experiment are shown in Fig.~\ref{fig:random_noise} using the CelebA.  Each column shows how the output changes as we change the noise vector.  Different subjects are shown in different rows.  It is interesting to note that, as we change the noise vector, attributes stay the same while the identity changes.  This can be clearly seen by comparing the synthesized results in each row.

\subsection{Quantitative Results}
In addition to the qualitative results presented in Fig.~\ref{fig:CelebA_result}, \ref{fig:LFW_result}, we present quantitative comparisons in Table \ref{tab: quantitative result}. Since the ground-truth images corresponding to the noise-generated images are not available, we choose the quantitative criterion based on the Fréchet Inception Distance (FID) \cite{heusel2017gans,lucic2018gans} and Attribute $L_{2}$-norm.
The FID is a measure of similarity between two datasets of images. It was shown to correlate well with human judgment of visual quality and is most often used to evaluate the quality of samples generated by GANs.  Attribute $L_{2}$-norm is used to compare the quality of attributes corresponding to different images. We extract the attributes from the synthesized images as well as the reference image using the MOON attribute prediction method \cite{rudd2016moon}.  Once the attributes are extracted, we simply take the $L_{2}$-norm of the difference between the attributes as follows
\begin{equation}
\text{Attribute  } L_{2}=\|\hat{a}_{ref}-\hat{a}_{synth}\|_{2},
\end{equation}
where $\hat{a}_{ref}$ and $\hat{a}_{synth}$ are the 23 extracted attributes from the reference image and the synthesized image, respectively.  Note that lower values of the FID score  and the Attribute $L_{2}$ measure imply the better performance.   The  quantitative results corresponding to different methods on the CalebA and LFW datasets are shown in Table~\ref{tab: quantitative result}.  Results are evaluated on the test splits of the corresponding dataset and the average performance along with the standard deviation are reported in Table~\ref{tab: quantitative result}.

As can be seen from this table, the proposed method produces the lowest FID scores implying that the images generated by our method are more realistic than the ones generated by other methods.  Furthermore, our method produces the lowest Attribute $L_{2}$ scores.  This implies that our method is able to generate attribute-preserved images better than the other compared methods.   This can be clearly seen by comparing the images synthesized by different methods in Fig.~\ref{fig:CelebA_result} and Fig.~\ref{fig:LFW_result}.

%\section{Limitations}
%In this work, we proposed a attribute-sketch and sketch-face framework for facial images synthesis from visual-attributes. This work achieves a better results when compared with several adopted text-to-image methods. However, when adopt our proposed method into high resolution image database, we find there is a main limitations of our proposed method: the GPU computation cost is high. This is because our proposed method needs two generators for synthesizing both high resolution sketch and facial images, which double the GPU memory requirement. In this work, we reduce the batch-size during training the network, but this makes the training time is very long. It takes more than 30 hours to finish the training procedure  on our proposed methods. 

\section{Conclusion}\label{sec:con}
We presented a novel deep generative framework for reconstructing face images from visual attributes. Our method makes use of an intermediate representation to generate photo realistic images.  The training part of our method consists of two models: Sketch Generator Network and Face Generator Network. Multi-scale  hierarchical network architectures are proposed for each generator networks.  Various experiments on three publicly available datasets show the significance of the proposed synthesis framework.  In addition, an ablation study was conducted to show the importance of different components of our network.    Various experiments showed that the proposed method is able to generate high-quality images and achieves significant improvements over the state-of-the-art methods.

On of the limitations of this work is that the synthesized images do not preserve the identity.  In the future, we will develop methods that can synthesize identity-preserving images from visual attributes.  These images can then be used to augment the datasets for face recognition \cite{wu2018light}, \cite{JC_WACV2016}.

\section*{Acknowledgment}

This research is based upon work supported by the Of- ficeof the Director of National Intelligence (ODNI), IntelligenceAdvanced Research Projects Activity (IARPA), via IARPA R\&D Contract No. 2019-19022600002. The views and conclu-sions contained herein are those of the au- thors and should notbe interpreted as necessarily represent- ing the official policiesor endorsements, either expressed or implied, of the ODNI,IARPA, or the U.S. Government. The U.S. Government is authorized to reproduce and dis- tribute reprints for Governmentalpurposes notwithstanding any copyright annotation thereon.

\begin{IEEEbiography}
	[{\includegraphics[width=1in,height=1.25in,clip,keepaspectratio]{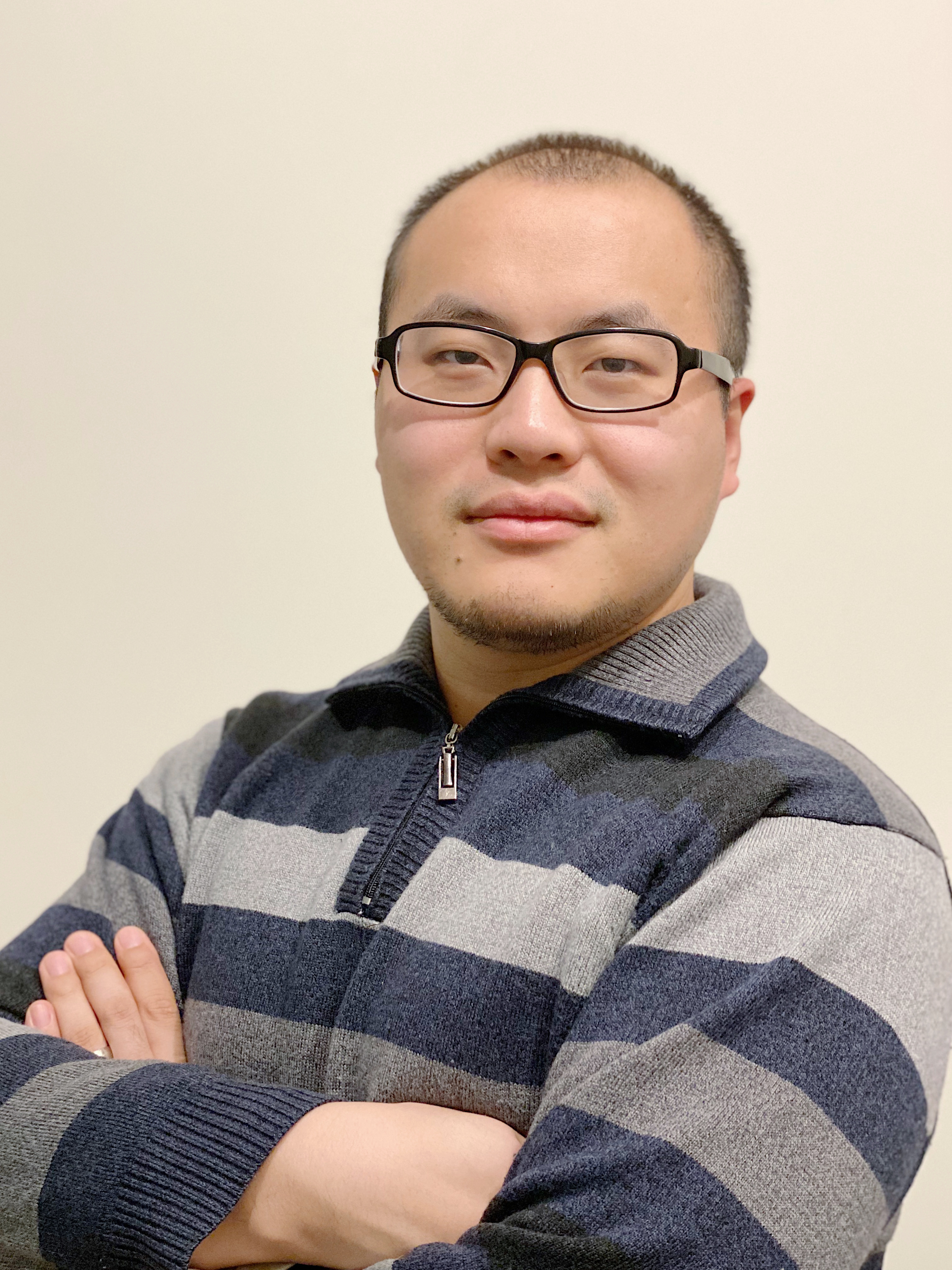}}]%
	{Xing Di}
	is a Ph.D. student in the Department of Electrical and Computer Engineering (ECE) at Johns Hopkins University. Prior to joining Hopkins, he was an Ph.D. student in the Department of ECE at Rutgers University. He completed his M.E. in Electrical Engineering from the Stevens Institute of Technology, Hoboken, NJ, in 2015. His current research interests include machine learning, computer vision and image process with  applications  in  biometrics.  He has received a Best Student Paper Awards at IAPR ICPR 2018. He has received the ECE student development award in Rutgers. He also serves as the journal reviewer in IEEE-TIP, PR, and conference reviewer in BTAS and ICB.
\end{IEEEbiography}

\begin{IEEEbiography}
	[{\includegraphics[width=1in,height=1.25in,clip,keepaspectratio]{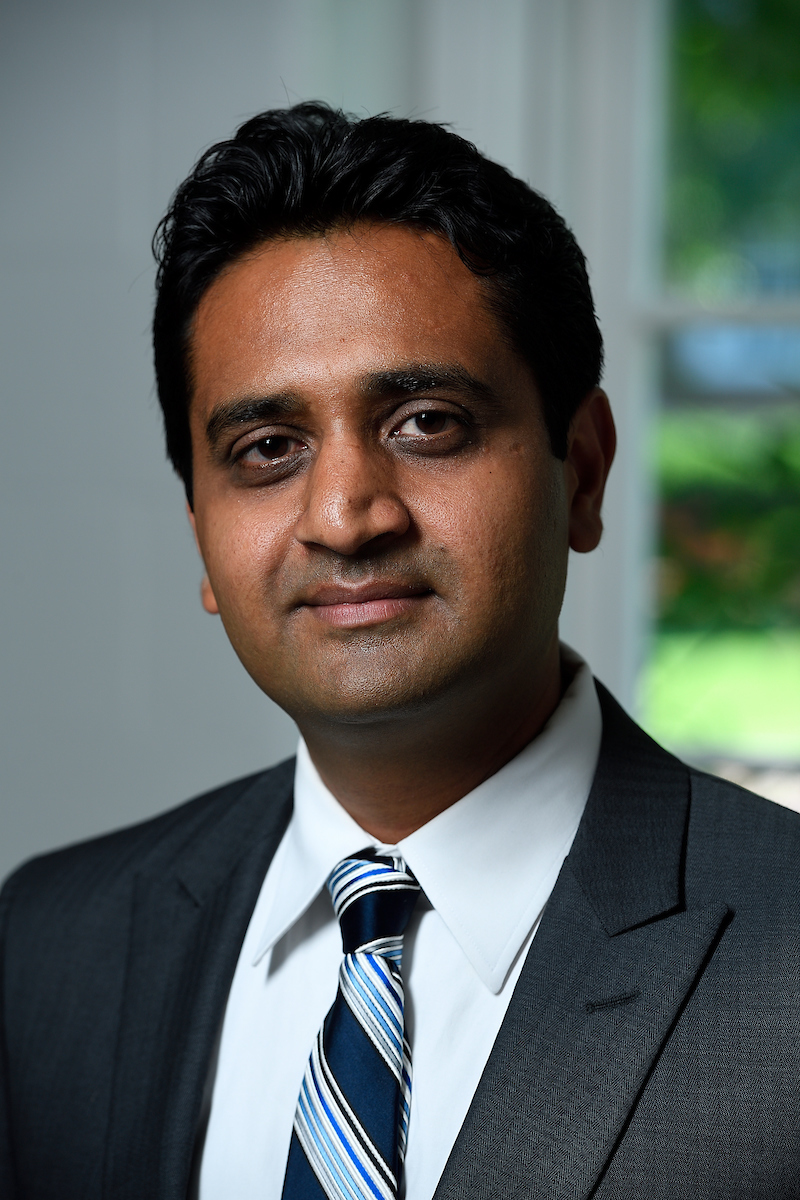}}]%
	{Vishal M. Patel}
	\text{[SM'15]} is an Assistant Professor in the Department of Electrical and Computer Engineering (ECE) at Johns Hopkins University. Prior to joining Hopkins, he was an A. Walter Tyson Assistant Professor in the Department of ECE at Rutgers University and a member of the research faculty at the University of Maryland Institute for Advanced Computer Studies (UMIACS). He completed his Ph.D. in Electrical Engineering from the University of Maryland, College Park, MD, in 2010. His current research interests include computer vision, image processing, and pattern  recognition  with  applications  in  biometrics  and  imaging.  He has received a number of awards including the 2016 ONR Young Investigator Award, the 2016 Jimmy Lin Award for Invention, A. Walter Tyson Assistant Professorship Award, Best Paper Award at IEEE AVSS 2017 \& 2019, Best Paper Award at IEEE BTAS 2015, Honorable Mention Paper Award at IAPR ICB 2018, two Best Student Paper Awards at IAPR ICPR 2018, and Best Poster Awards at BTAS 2015 and 2016. He is an Associate Editor of the IEEE Signal Processing Magazine, IEEE Biometrics Compendium, and serves on the Information Forensics and Security Technical Committee of the IEEE Signal Processing Society. He is a member of Eta Kappa Nu, Pi Mu Epsilon, and Phi Beta Kappa.
\end{IEEEbiography}

\bibliographystyle{IEEEtran}
\bibliography{Attr2Sk2Face.bib}
\end{document}